\documentclass[letterpaper, 10 pt, conference]{ieeeconf}

\IEEEoverridecommandlockouts

\usepackage{times}
\usepackage{amsmath,amsfonts,amssymb}
\usepackage[ruled,vlined]{algorithm2e}
\usepackage{graphicx}
\usepackage{mdframed}
\usepackage{subfigure}
\usepackage{todonotes}
\usepackage{multirow}
\usepackage{xcolor}
\usepackage[noadjust]{cite}
\usepackage{hyperref}
\usepackage{makecell}
\usepackage{balance}

\newcommand{\bi}[1]{\boldsymbol{#1}}

\title{Predictive Modeling of Periodic Behavior for\\ Human-Robot Symbiotic Walking}

\author{Geoffrey Clark$^{1}$, Joseph Campbell$^{1}$, Seyed Mostafa Rezayat Sorkhabadi$^{2}$, Wenlong Zhang$^{2}$, Heni Ben Amor$^{1}$%
\thanks{$^{1}$G.~Clark, J.~Campbell, and H.~Ben~Amor are with the School of Computing, Informatics, and Decision Systems Engineering, Arizona State University
        {\tt\small \{gmclark1, jacampb1, hbenamor\}@asu.edu}}%
\thanks{$^{2}$S.M.R.~Sorkhabadi and W.~Zhang are with The Polytechnic School, Arizona State University, Mesa, AZ, 85212, USA. 
        {\tt\small \{srezayat, wenlong.zhang\}@asu.edu}}%
\thanks{This work was funded by the National Science Foundation under the Career Award grant number: FP00012258}
}

\begin{document}

\maketitle

\begin{abstract}
 We propose in this paper Periodic Interaction Primitives -- a probabilistic framework that can be used to learn compact models of periodic behavior. Our approach extends existing formulations of Interaction Primitives to periodic movement regimes, i.e., walking. We show that this model is particularly well-suited for learning data-driven, customized models of human walking, which can then be used for generating predictions over future states or for inferring latent, biomechanical variables. We also demonstrate how the same framework can be used to learn controllers for a robotic prosthesis using an imitation learning approach. Results in experiments with human participants indicate that Periodic Interaction Primitives efficiently generate  predictions and ankle angle control signals for a robotic prosthetic ankle, with MAE of 2.21$^\circ$ in 0.0008s per inference. Performance degrades gracefully in the presence of noise or sensor fall outs. Compared to alternatives, this algorithm functions 20 times faster and performed 4.5 times more accurately on test subjects.
\end{abstract}

\section{Introduction}
Walking is a critical motor skill which is at the center of human mobility and independence.
Healthy human adults on average walk several thousand steps per day seemingly without any effort and with substantial grace and fluency to their movements.
However, for many millions of people~\cite{ziegler2008estimating} affected by musculoskeletal disorders, amputations, neurologic pathologies, or other health-related issues walking is a daily struggle or even completely out of reach~\cite{medhat1990factors,buzgova2009}.
Modern assistive robotics technology (e.g. an exoskeleton, orthotic or prosthetic) has the potential to change the lives of  people affected by such conditions for the better, by replacing missing, or augmenting existing capabilities.
However, methodologies are needed that allow robots to generate periodic actions that seamlessly blend with those of the human user~\cite{weir2003great}.

In particular, such assistive robots need to be able to anticipate future kinematic states of the human partner given current sensor readings of their walking gait \cite{klute2009lower}, thereby providing a window of opportunity for decision-making and control.
In addition to predicting kinematics, it is suggested that assistive devices also take biomechanical and ergonomic ramifications on the human body into account~\cite{morgenroth2011effect}.
Considering that different walking gaits result in varying magnitudes in internal stresses being applied to the human musculoskeletal system; it is important that assistive robots can generate estimates of biomechanical variables in a rapid and predictive fashion to avoid overexertion, injury, or even serious chronic diseases such as osteoarthritis (OA)~\cite{MORGENROTH2012S20}.
Inverse dynamics~\cite{robertson2013research} techniques can generate accurate estimates of internal biomechanical variables, but are generally slow and non-predictive in nature. 
In addition, they often require information from high-fidelity sensors, usually motion capture and an instrumented treadmill with force plates, which is typically not available outside of controlled laboratory environments.
Hence, other methods are needed to be used in non-clinical settings with low-fidelity sensors and can run on devices with limited computational power.

\begin{figure}[t!]
    \vspace{-0.25cm}
	\begin{center}
		\includegraphics[width=0.7\columnwidth]{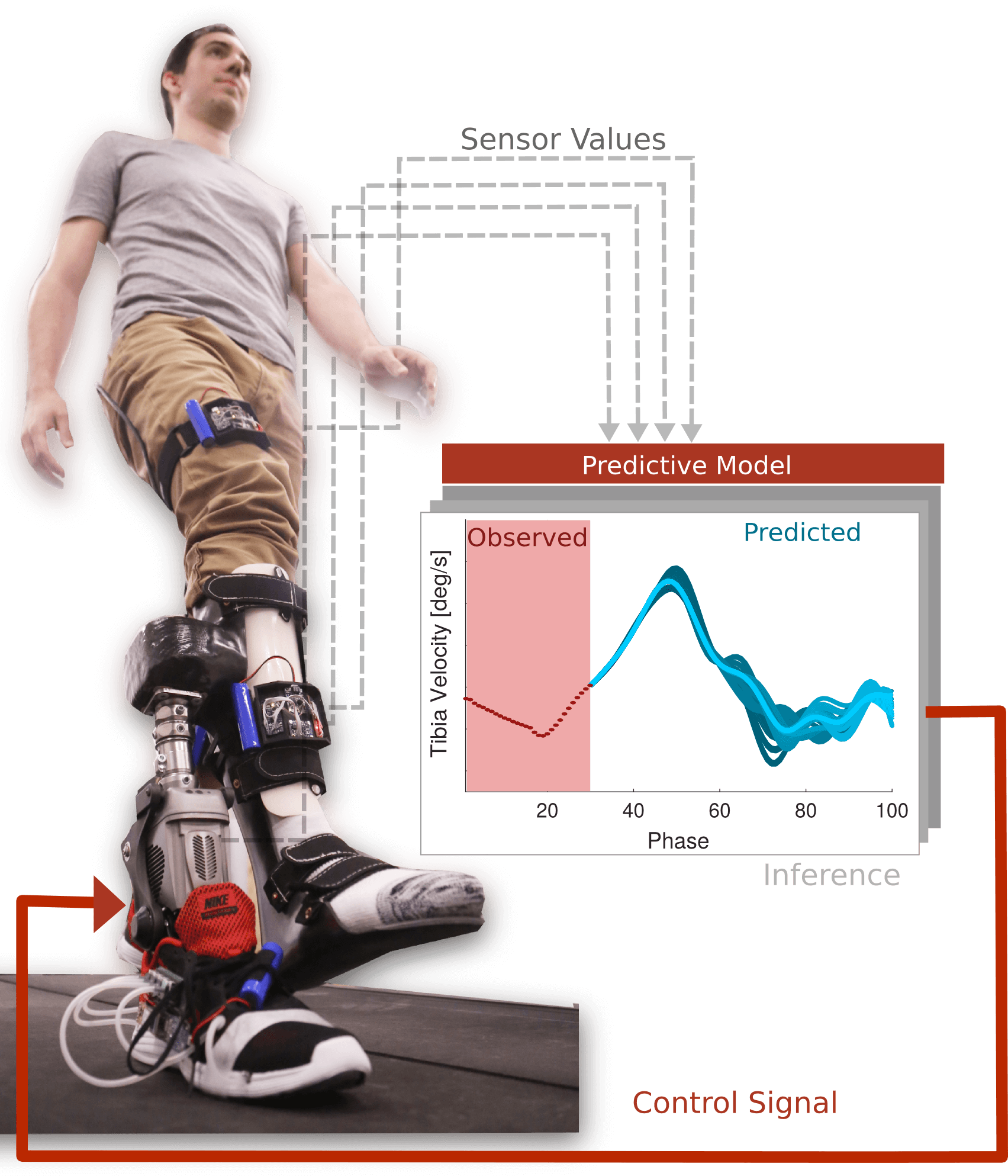}
	\end{center}
	\vspace{-0.5cm}
	\caption{PIP is used to learn a realtime, closed-loop controller for symbiotic walking by analyzing and predicting sensor values, as well as control signals.}
	\vspace{-.75cm}
	\label{fig:teaser}
\end{figure}

Therefore we propose Periodic Interaction Primitives (PIP) -- an extension to interaction primitives (IP)~\cite{amor2014interaction}.
PIP is a data-driven approach to the predictive modeling of periodic behavior for human-robot symbiotic walking, i.e., effortless walking with an intelligent prosthesis, exoskeleton, or other assistive device.
Once trained, a PIP can be used for anticipating the future states of the human user, such as kinematics from observed sensors, and inferring latent, unobserved variables, for instance internal biomechanical variables or external prosthesis or exoskeleton control without the further necessity of a motion capture system or additional biomechanical model.
These features are achieved, by correlating at runtime, the information captured from multiple sensors with corresponding biomechanical or control parameters which are incorporated during the training process.
PIP is differentiated from previous interaction primitives in three ways.
(1) We tie the statistical processes of IP to a periodic domain through the use of von Mises basis functions and provide analytical formulations of partial derivatives with respect to phase.
(2) In concert with periodicity modifications, PIP incorporates a novel phase estimation method which is effective in tightly-coupled or symbiotic systems which exhibit periodic phase behavior.
(3) We describe a data augmentation scheme that allows PIPs to infer unobservable, biomechanical information through latent variable estimation.

\section{Related Work}
Early work on control involving prosthetics and exoskeletons has primarily focused on the control aspects of the human robot system~\cite{tucker2015control}.
Controllers have been successfully developed for both transtibial~\cite{au2009powered} and transfemoral~\cite{sup2009preliminary} applications.
Extensions have also been made,
which provide control parameters for a variety of common terrains such as navigating slopes~\cite{sup2010upslope} or stairs ~\cite{au2008powered}.
While these methods have been shown to produce robust and stable controllers, there are a number of downsides.
First, these control systems are entirely reactive to the human subject.
Second, they must be closely tuned to match each individuals walking gait.

In contrast, data-driven approaches to the probabilistic modeling of HRI often leverage hidden Markov models~\cite{rabiner1989, donghuilee2010, Rozo16Frontiers},
which are well-suited for the joint inference problems commonly encountered in interaction.
However, they often require a discretization of the state space and suffer from computational overhead too large to be feasibly placed onto the low-power microprocessors that prosthetic devices require.

One method that overcomes some of these limitations is called Probabilistic Movement Primitives (ProMP)~\cite{paraschos2013probabiistic,dermy2017prediction}.
As a probabilistic formulation of the Movement Primitivies concept, this method uses a stochastic feedback controller to reproduce learned trajectory distributions, both stroke based and rhythmic, with predefined phase progression.
By incorporating a probabilistic framework (utilizing the demonstration variance) ProMPs allow for greater flexibility in trajectory reproduction than deterministic methods.

In order to increase flexibility and accuracy even further Interaction Primitives were proposed, as a learning from demonstration method in which a joint probability distribution is created over the parameterized models of two interacting agents.
By incorporating the full joint trajectory distribution IPs enable the inference of all modeled parameters based on a sample set of observed parameters.
This approach has proven capable in several different scenarios~\cite{Cui2018, campbell2017bayesian, ogur2018, chen2017, ewerton2015}.
However while potential for rhythmic behaviors has been proposed before, careful management of basis functions as well as modifications to phase estimation must be made to to take periodicity into account.
Additionally, these methods do not yet address the problem of inferring temporal position in highly dynamic tightly coupled periodic systems; where the behavior of the human influences the behavior of the robot and vice versa.
Instead they utilize methods such as dynamic time warping (DTW) to help determine the correct phase of the system for interaction, which are too computationally demanding  to be used in run in real-time on assistive devices.
This type of symbiosis between the human and robot, where there is constant reaction and interaction, necessitates new methods of quantifying and modeling interactions.

\section{Predictive Modeling with Periodic Interaction Primitives}
The approach proposed in this paper focuses on probabilistic modeling and inference of periodic behavior during walking. 
The main goal is to learn a predictive model that relates current multimodal sensor readings to future states of the human during the walking gait. 
In contrast to previous modeling approaches, we simultaneously predict: (a) observable sensor values, (b) unobservable biomechanical data, as well as potentially (c) control values for a prosthesis, exoskeleton or other assistive device.

As an overview of the algorithm presented in this paper; first a data set is collected which represents typical variations in sensor readings during walking. 
The data set is then augmented with biomechanical information from the subject via tools such as OpenSim~\cite{delp2007opensim} or Vicon Nexus~\cite{vicon}, as well as control signals for which action an assistive device or prosthesis should take at each timestep.
The final step of the learning process is to train the predictive model, which extracts and efficiently represents the correlations and dynamics inherent in the data.
The goal of PIP is to represent the reciprocal coupling between observed data (sensor values), unobservable data (extracted biomechanical parameters), and control variables (angular control value for a prosthetic ankle), instead of an analytical formulation of the dynamics.

\subsection{Training: Data Collection and Data Augmentation}
\label{sec:augmentation}
One benefit of PIP is that variables which are not observed though the sensors such as: joint torques, muscle lengths, metabolic power consumption, or mechanical work can be inferred.
We accomplish this by incorporating the internal biomechanical variables and their distributions w.r.t. time directly into our predictive models; in the following manner.

Each user performs a walking action spanning multiple gait cycles which produces a time series of sensor measurements.
This time series is subdivided into a set of $N$ individual demonstrations, each representing a single gait cycle, $[ \bi{y}_1, \dots ,\bi{y}_{T_n} ] \in \mathbb{R}^{D_s \times T_n}$, such that $D_s$ is the total number of degrees of freedom from all sensors and $T_n$ is the number of observations for the $n$-th demonstration.

Given the set of individual demonstrations, the trajectories of unobservable variables $[\bi{m}_1, \dots ,\bi{m}_{T_n}] \in \mathbb{R}^{D_m \times T_n}$, such that $D_m$ is total number of degrees of freedom for all unobservable variables, is calculated using a desired analysis tool.
The original sensor observations are then augmented such that $\boldsymbol{Y}_{1:t} = [\bi{y}_t, \bi{m}_t] \in \mathbb{R}^{D}$ forms the new state representation of the human at every time step, where $D = D_s + D_m$.

The objective of this approach is to turn the estimation of internal variables into a function approximation problem which will run significantly faster
and will work even when variables are missing.  
By incorporating unobservable variables into the training process, we can later leverage the PIP framework to infer internal loads using observable variables alone.

\subsection{Training: Learning a Periodic Interaction Primitive}
\label{sec: promp for interaction}
Probabilistically, the objective of PIP is to infer estimates of the future states of both the observed and non-observed variables $\hat{\boldsymbol{Y}}_{t+1:T}$, given observations of the human $\boldsymbol{Y}_{1:t}$:
\begin{align}
    p(\hat{\boldsymbol{Y}}_{t+1:T} | \boldsymbol{Y}_{1:t}).
    \label{eq:pip_general1}
\end{align}
This requires us to define a state transition model for each degree of freedom along with an observation model relating the unobservable variables to the observed variables based on our set of training demonstrations.
As previously established~\cite{campbell2017bayesian}, we can simplify this process significantly by first transforming our state representation into a latent space via basis function decomposition.
That is, we want to find a vector of coefficients $\boldsymbol{w}^d$ for each degree of freedom  $0 \leq d < D$ such that $y^d_{t} = \Phi_{\phi(t)}^{\intercal} \boldsymbol{w}^d + \epsilon_y$ for all observations in the training demonstrations, where $\Phi_{\phi(t)} \in \mathbb{R}^{B^d \times 1}$ is a vector of $B^d$ basis functions, $\boldsymbol{w}^d \in \mathbb{R}^{B^d \times 1}$, and $\epsilon_y$ is i.i.d. Gaussian noise.

This decomposition accomplishes three goals: a) it produces a time-invariant state $\boldsymbol{w}^d$ for each DOF, allowing inference of all past and future values from a single state value, b) it captures the dynamics of each DOF over time, eliminating the need for an explicit transition function, and c) it allows for correlation between different DOFs.
A necessary component of the above decomposition is the introduction of a temporal dependence for the basis functions on a relative phase function $\phi(t) \in [0, 100]$, or for notational simplicity $\phi$.
Utilizing phase decouples the relative progress of an interaction from its absolute length, thus preserving the shape of a trajectory across temporal speeds and allows for the efficient calculation of the values for a DOF for an entire interaction: $\boldsymbol{Y}^d_{1:P} = [\boldsymbol{\Phi}^{\intercal} \boldsymbol{w}^d]$ where $\boldsymbol{\Phi} = [\Phi_{0}, \dots, \Phi_{100}] \in \mathbb{R}^{B \times P}$.
$P$ here represents the number of sampled points for the trajectory and has the additional benefit of acting as a smoothing function.
We succinctly represent all DOFs in our state representation by concatenating them into a single weight vector: $\boldsymbol{w} = [\boldsymbol{w}^{0\intercal}, \dots, \boldsymbol{w}^{D\intercal}] \in \mathbb{R}^{1 \times B}$ with $B = \sum_d^D B^d$.

By its very nature walking is a periodic action and biomechanical analyses have therefore shown that gait cycles can be quantitatively described by phase plane evaluations~\cite{mcmillan2010sagittal}.
So as to truly capture the periodic nature of our state variables, we must choose an appropriate basis function which operates over a cyclical domain.
This is in contrast to prior work in IPs which employ Gaussian basis functions for trajectories with distinct start and end points.
In this work, we utilize the von Mises function, an approximation of the wrapped normal distribution over the domain of the unit circle:
\begin{align}
    \psi_{\mu}(\phi) &= \frac{e^{\kappa cos(\alpha (\phi-\mu))}}{2\pi I_0(\kappa)},
\end{align}
where $\mu$ lies in the interval $[0, 2\pi]$ and specifies the center of the basis function, $\alpha$ is $2\pi/100$ (required to transform the phase into $[0, 2\pi]$), $\kappa$ is a measure of inverse variance, and $I_0$ is the modified Bessel function of order zero.

\subsection{Inference: Phase Detection}
With the basis space fully defined, we re-formulate our probabilistic objective from~(\ref{eq:pip_general1}) as
\begin{equation}
\label{eq:pip_general2}
p(\boldsymbol{w}_t | \boldsymbol{Y}_{1:t}, \boldsymbol{w}_{0}) \propto p(\boldsymbol{y}_{t} | \boldsymbol{w}_t) p(\boldsymbol{w}_t | \boldsymbol{Y}_{1:t-1}, \boldsymbol{w}_{0}).
\end{equation}
Intuitively, the goal is to infer the state of the interaction $\boldsymbol{w}_t$ given $t$ observations of the human and a prior estimate of the state, $\boldsymbol{w}_0$.
However, without knowing the phase values associated with the observations $\boldsymbol{Y}_{1:t}$ we cannot apply a correction to our estimate, and so we must first estimate the phase $\phi$ for each observation $\boldsymbol{y}_t$.
As part of the training process we create a low dimensional manifold, or subspace, encompassing the phase and observed variables. The created manifold allows for the efficient projection of observations onto the manifold so as to determine the corresponding phase.

Before we can construct this manifold model, we must first temporally align the demonstration gait cycles. This is required as gaits are nonlinearly warped in time through variations in walking speed, incline, and other environmental and biomechanical factors.
According to existing analyses on complex joint movements, there are quantitative relationships between the positions of the knee, tibia, and ankle with respect to the phase plane. In this work, we exploit these relationships to temporally align our demonstrations through measurements of the angular position, angular velocity, and angular acceleration of the tibia. While the position and velocity are directly observed, the acceleration is not.
Instead it is computed through the partial derivative of the linear combination of basis functions with respect to the phase:
\begin{align}
    \frac{\partial \sum_b^{B^d} {\psi_{b}(\phi) w^d_b}}{\partial \phi} &= %
    \sum_b^{B^d} w^d_b \frac{\partial \psi_{b}(\phi)}{\partial \phi} \\
    &= \sum_b^{B^d} w^d_b \frac{\kappa \sin(\alpha(\mu_b - \phi))e^{\kappa \cos(\alpha(\mu_b - \phi))}}{2 \pi I_o(\kappa)}, \nonumber
\end{align}
where $d$ corresponds to the tibia angular velocity DOF.
Temporal alignment is performed with dynamic time warping (DTW)~\cite{Mueller2007}, in which the distance between two demonstrations $u$ and $v$ is computed as a cost matrix $C$.
The optimal alignment can be found as the minimum path through the cost matrix,
\begin{align}
    C_{p^*}(u,v) &= min\Big\{ \sum_{l=1}^L  c(u_{n_l},v_{m_l})\Big\},
\end{align}
where the cost function $c(\cdot)$ is the sum of the Euclidean distances between angular positions, velocities, and accelerations. 
Since we wish to align all demonstrated trajectories, the full set of trajectories are temporally aligned creating an approximately optimal time alignment between all demonstrations.
We assume that all aligned demonstrations lie over the phase interval $[0, 100]$, such that the first time step is $0$ and the last is $100$.
Finally, we discretize the data over two dimensions -- $E$ angular position states and $F$ angular velocity states -- and assign each discrete state a phase value determined via the nearest neighbor in the set of aligned trajectories.
This creates an $E \times F$ lookup table, in the subspace or manifold, that can be queried quickly to approximate phase from angular position and velocity.
We denote the lookup function $L(\boldsymbol{y}_t)$ which accepts an observation from which the angular position and velocity are projected onto the table and the resulting phase $\phi$ is returned.

\subsection{Inference: Generating Predictions and Controls}
\label{sec:inference}
In order to evaluate the posterior probability of (\ref{eq:pip_general2}), we utilize a full state linear estimator, i.e. Kalman filter, to recursively apply Gaussian conditioning to a matrix of basis weights.
Which evaluates the state prediction density as,
\begin{align}
\label{eq:bip_prediction_reduced}
& p(\boldsymbol{w}_t | \boldsymbol{Y}_{1:t-1}, \boldsymbol{w}_{0}) \nonumber \\
& = \int p(\boldsymbol{w}_t | \boldsymbol{w}_{t-1}) %
p(\boldsymbol{w}_{t-1} | \boldsymbol{Y}_{1:t-1}, \boldsymbol{w}_{0})d\boldsymbol{w}_{t-1}
\end{align}
For tractability, we make the assumption that the posterior state densities are normally distributed, that is, $p(\boldsymbol{w}_t | \boldsymbol{Y}_{1:t}, \boldsymbol{w}_{0}) = \mathcal{N}( \boldsymbol{\mu}_{t}, \boldsymbol{\Sigma}_{t})$.
Additionally, because our state has been transformed into a time-invariant representation, we no longer have a meaningful state transition function in time.
That is, $p(\boldsymbol{w}_{t} | \boldsymbol{Y}_{1:t-1}, \boldsymbol{w}_{0}) = p(\boldsymbol{w}_{t-1} | \boldsymbol{Y}_{1:t-1}, \boldsymbol{w}_{0})$.
The observation function, $h(\cdot)$, is simply the linear combination of weighted basis functions as described in Sec.~\ref{sec: promp for interaction} and the observation matrix $\boldsymbol{H}$ is defined as:
\begin{align}
    \bi{H}_t=
	\begin{bmatrix}
    \Phi_{\phi} & \dots & 0\\
    \vdots & \ddots & \vdots\\
    0 & \dots & \Phi_{\phi}
    \end{bmatrix}.
\end{align}
The standard update equations for the calculation of the posterior density follow:
\begin{align}
    \bi{K}_{t} &= \bi{\Sigma}_{t-1} \bi{H}_t^{\intercal} (\bi{H}_t \bi{\Sigma}_{t-1} \bi{H}_t^{\intercal} + \bi{R}_t)^{-1},\\
    \bi{\mu}_{t} &= \bi{\mu}_{t-1} + \bi{K}_t(\bi{y}_t - \bi{H}_t \bi{\mu}_{t-1}),\\
    \bi{\Sigma}_{t} &= (I - \bi{K}_t \bi{H}_t)\bi{\Sigma}_{t-1},
\end{align}
where $\boldsymbol{K}$ is the Kalman gain matrix which controls how heavily we update our state estimate based on the observed measurement while taking into account the measurement noise matrix $\boldsymbol{R}_t$.
While the observation matrix given here includes both the observed and controlled DOFs, in practice only the observed DOFs are considered.
This can be orchestrated by either setting the entries corresponding to the controlled DOFs to zero in $\boldsymbol{H}_t$, or regularizing them with artificial noise in $\boldsymbol{R}_t$.
Future trajectories for any degree of freedom $d$, including the controlled ones, can be trivially calculated with $\boldsymbol{Y}^d_{1:P} = [\boldsymbol{\Phi}^{\intercal} \boldsymbol{\mu}_t^d]$.
The initial estimate, $p(\boldsymbol{w}_0) = \mathcal{N}( \boldsymbol{\mu}_0, \boldsymbol{\Sigma}_0)$, is determined from the set of basis weights generated from the demonstrations, where $\boldsymbol{\mu}_0$ is the sample mean and $\boldsymbol{\Sigma}_0$ is the sample covariance.

The PIP algorithm is shown in its entirety in Fig.~\ref{fig:algorithm_1}.
At the first time step, the previous state estimate is given as $\boldsymbol{\mu}_0$ and $\boldsymbol{\Sigma}_0$.
In subsequent steps, the updated state estimate is used as the prior.

\begin{figure}[t!]
\vspace{.2cm}
\begin{mdframed}
\textbf{Periodic Interaction Primitives}
\vspace{0.7em}

\textbf{Input: } $L(\cdot)$: phase lookup function, $\bi{\mu}_{t-1} \in \mathbb{R}^{1 \times B}$: prior distribution mean, $\bi{\Sigma}_{t-1} \in \mathbb{R}^{B \times B}$: prior distribution uncertainty, $\bi{y}_t \in \mathbb{R}^{1 \times D_s}$: sensor observations at $t$.
\vspace{0.2em}

\textbf{Output: } $\hat{\bi{Y}}^{D_s}_{1:P} \in \mathbb{R}^{P \times D_s}$: inferred states of $D_s$ observed variables, $\hat{\bi{Y}}^{D_m}_{1:P} \in \mathbb{R}^{P \times D_m}$: inferred states of $D_m$ unobservable variables, $\boldsymbol{\mu}_t$: the updated state mean, $\boldsymbol{\Sigma}_t$: the updated state uncertainty.
\vspace{0.7em}
\hrule
\vspace{0.7em}
\begin{enumerate}
    \item Estimate phase $\phi$ using trained lookup function:
    \begin{align}
        \phi = L(\boldsymbol{y}_t). \nonumber
    \end{align}
    \item Compute the Kalman gain matrix:
    \begin{align}
    \bi{K}_{t} &= \bi{\Sigma}_{t-1} \bi{H}_t^{\intercal} (\bi{H}_t \bi{\Sigma}_{t-1} \bi{H}_t^{\intercal} + \bi{R})^{-1}. \nonumber
    \end{align}
    \item Incorporate observations into the state estimate:
    \begin{align}
    \bi{\mu}_{t} &= \bi{\mu}_{t-1} + \bi{K}_t(\bi{y}_t - \bi{H}_t \bi{\mu}_{t-1}), \nonumber\\
    \bi{\Sigma}_{t} &= (I - \bi{K}_t \bi{H}_t)\bi{\Sigma}_{t-1}. \nonumber
    \end{align}
    \item Predict the past, present, and future states of all observed and unobserved state variables:
    \begin{align}
        \hat{\boldsymbol{Y}}^d_{1:P} = [\boldsymbol{\Phi}^{\intercal} \boldsymbol{w}^d] \quad \text{for} \quad 0 < d < D. \nonumber
    \end{align}
    \item \textbf{Output} the observed states, unobservable states, and posterior distribution:
    \begin{align}
        \hat{\boldsymbol{Y}}^{D_s}_{1:P}, \hat{\boldsymbol{Y}}^{D_m}_{1:P}, \boldsymbol{\mu}_t,  \boldsymbol{\Sigma}_t. \nonumber
    \end{align}
\end{enumerate}
\end{mdframed}
    \vspace{-.25cm}
    \caption{Periodic Interaction Primitives}
    \vspace{-.5cm}
    \label{fig:algorithm_1}
\end{figure}

\section{Experimental Setup}\label{sec:experiments}

\subsection{Data Collection}
Data collection was performed in a human subject study approved by the Institutional Review Board at Arizona State University.
Five participants were fitted with smart shoes, IMUs and retroreflective markers for a VICON motion capture system with 10 cameras.
Four IMU devices, one on each shank and one on each femur output 9DOF inertial data as well as angular position from a proprietary embedded sensor fusion algorithm.
The smart shoes measure ground reaction forces at four points: heel, first metatarsal joint, fourth metatarsal joint and toe, through silicone tubes which are wound into air bladders, placed in the soles of the shoes, and connected to barometric pressure sensors.
To collect biomechanical data, each participant was asked to walk on an instrumented treadmill at 5 different speeds from 0.5m/s to 1.3m/s for 2 min at each speed in a single trial.
Motion Capture marker positions and ground reaction force data was collected on the subjects and processed with the common commercially available biomechanics model: Vicon Plug‐In‐Gait model (version 2) in Vicon Nexus software~\cite{vicon} to calculate joint angles (angles between skeletal links), forces (reflected force from one skeletal link to another), and moments (torque between skeletal links). 
The entire learning process from raw data to testable model output takes approximately 3 minutes.
Subjects were divided into two groups three subjects for training and two for testing.
Ten consecutive strides from each subject were removed from the full data set as holdout data for testing and model evaluation.

\subsection{\textbf{Experiment 1}: Predictive Capabilities}
The first experiment evaluates the inference and prediction capabilities of the introduced PIP method by assessing: (a) the accuracy of the estimated phase variable when the subjects are walking at different speeds, and (b) the prediction quality of the learned models on observed variables.
\begin{figure}[ht!]
    \vspace{-.25cm}
	\begin{center}
		\includegraphics[width=1\columnwidth]{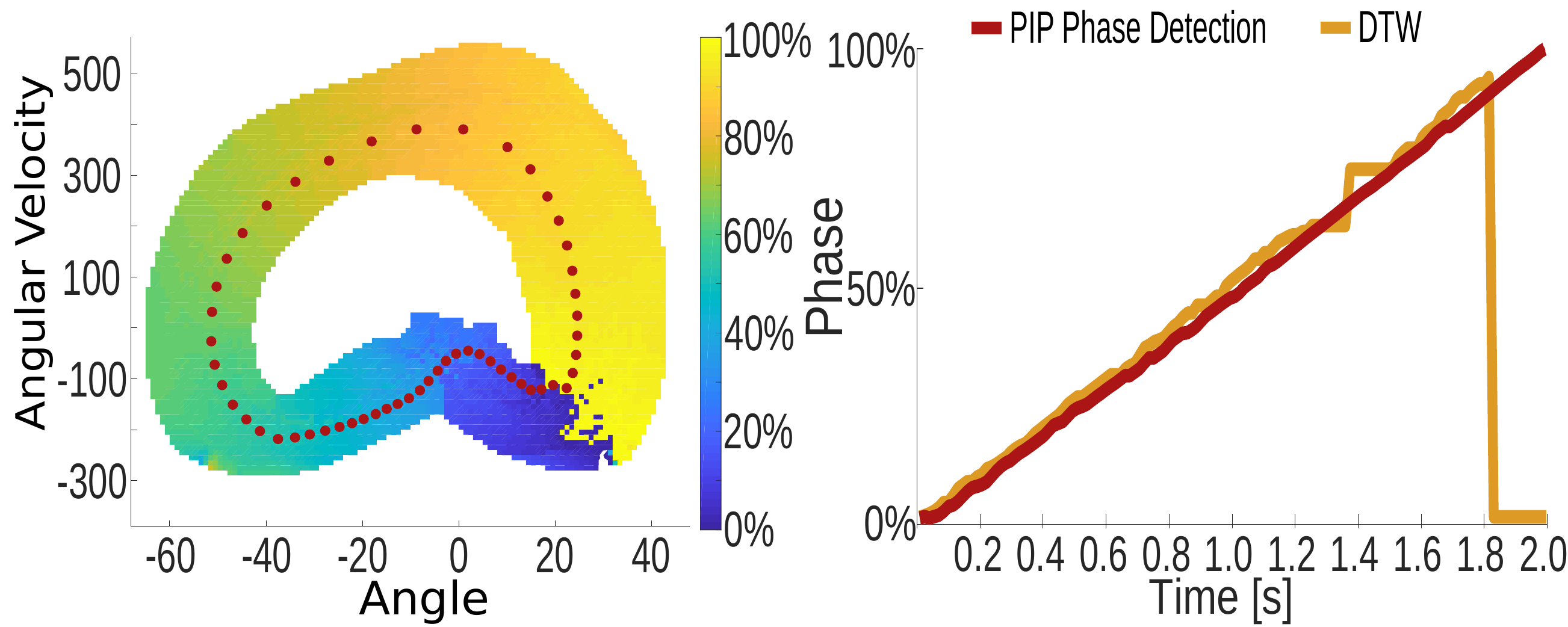}
	\end{center}
	\vspace{-.5cm}
	\caption{Left: Phase manifold, with individual observations (red points) used to generate phase estimate. Right: Phase as a function of time, with PIP phase detection method (red) and Dynamic Time Warping method (yellow).}
	\vspace{-.25cm}
	\label{fig:subjects_phase}
\end{figure}

\begin{figure}[ht!]
    \vspace{-.50cm}
	\begin{center}
		\includegraphics[width=1\columnwidth]{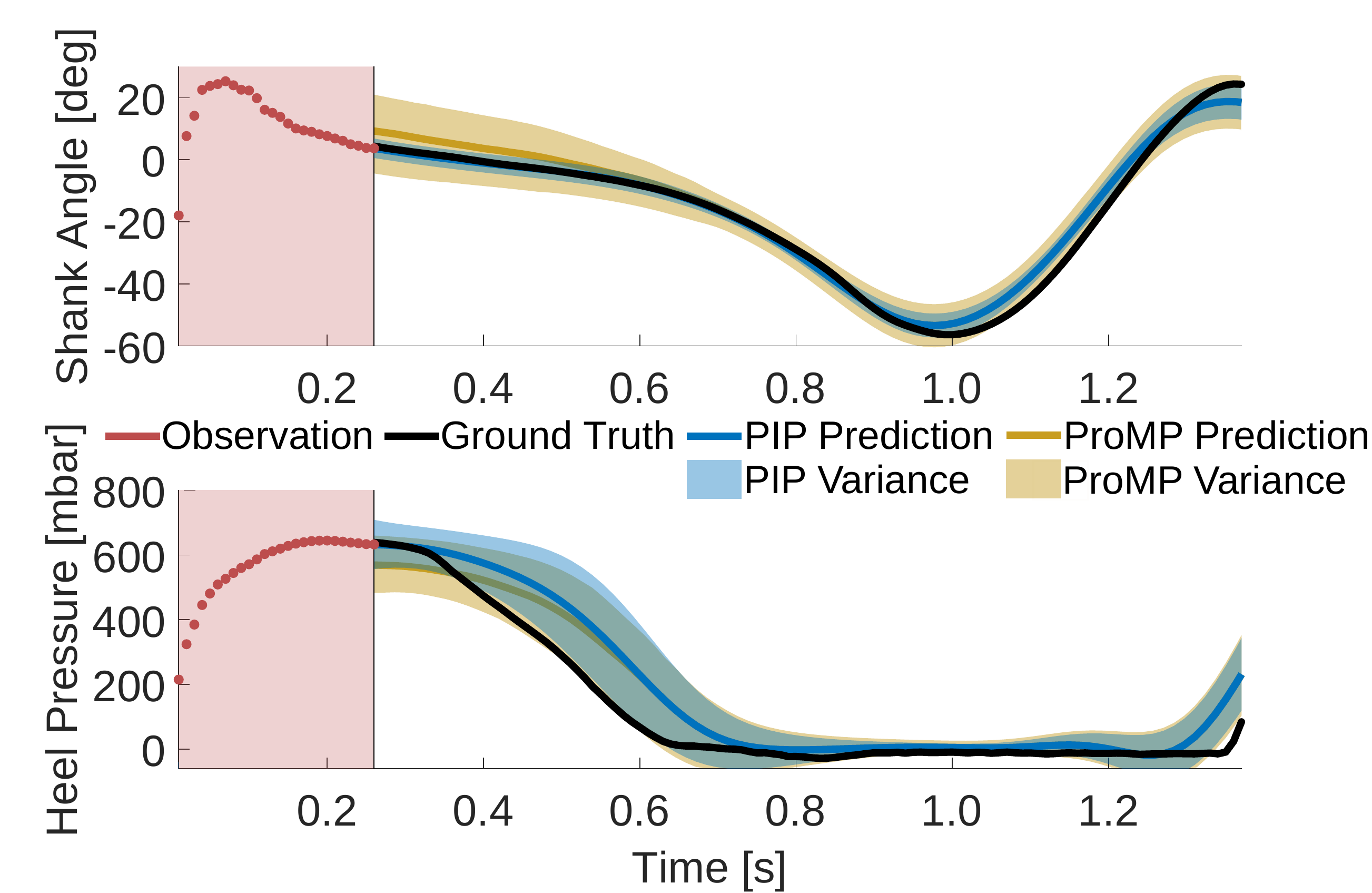}
	\end{center}
	\vspace{-.5cm}
	\caption{The red area corresponds to sensor observations (red points), with Predictions generated via PIP inference (blue), and prediction generated via ProMP with dynamic time warping (yellow).}
	\label{fig:subjects_prediction}
	\vspace{-.25cm}
\end{figure}
Fig.~\ref{fig:subjects_phase} shows an example of the PIP phase detection method as compared to the common method of dynamic time warping, with the phase manifold on the left plot and the detected phase on the right plot.
The phase manifold illustrates how each point in the space is directly associated with a specific phase value, when individual observations
occur, phase is directly estimated through the use of a 2D lookup function.
This leads to a very accurate phase detection which works over a large range of walking speeds.
The common phase detection method of Dynamic Time Warping is further compared against the PIP phase detection method to show that the PIP phase detection method is more accurate as it lacks the temporal disturbances commonly present in DTW.
During this experiment the inference time of each phase detection method was collected and it was found that the PIP method is significantly faster requiring $0.0004$ seconds average vs $0.008$ seconds average for DTW.

Table~\ref{tab:prediction_table} shows the mean absolute error (MAE) for prediction of sensor values via a PIP in the top section, using the joint model from all training subjects data, tested on each subject individually. We can see for example that the error on predicting the shank is in the range $1^{\circ}$ to $3^{\circ}$, hence yielding accurate predictions for all participants. The shank velocity is at approximately $15^{\circ}$. This may appear substantial, however, the prediction error amounts to a deviation of approximately $1.5\%$. In addition, all predictions of the foot pressure are lower than $30$ mBar and are therefore also of high accuracy.

\begin{table}[]
\vspace{.25cm}
    \centering
    \begin{tabular}{lcccc}
    \multicolumn{5}{c}{\textbf{Predicted Variables MAE}}\\
    \hline \hline
         & & Subject 1 & Subject 2 & Subject 3 \\
    \hline \hline
    Shank Angle & deg & 2.61 & 1.79 & 2.53 \\
    Shank Velocity & deg/s & 14.25 & 12.33 & 13.87 \\
    Femur Angle & deg & 1.85 & 1.11 & 1.83 \\
    Femur Velocity & deg/s & 6.13 & 6.03 & 7.72 \\
    Heel Pressure & mbar & 26.32 & 20.49 & 20.63 \\
    Meta 1 Pressure & mbar & 4.36 & 15.31 & 14.61 \\
    Meta 4 Pressure & mbar & 13.94 & 23.83 & 16.66 \\
    Toe Pressure & mbar & 14.56 & 28.84 & 26.41 \\
    \multicolumn{5}{c}{}\\
    \multicolumn{5}{c}{\textbf{Infered Variables MAE}}\\
    \hline \hline
         & & Subject 1 & Subject 2 & Subject 3 \\
    \hline \hline
    Ankle Angle & deg & 1.56 & 1.39 & 1.08 \\
    Ankle Moment & N.mm/Kg & 10.93 & 15.76 & 13.95 \\
    Ankle Force & N/Kg & 0.14 & 0.14 & 0.18 \\
    Knee Angle & deg & 1.95 & 1.91 & 1.30 \\
    Knee Moment & N.mm/Kg & 40.15 & 54.79 & 34.47 \\
    Knee Force & N/Kg & 0.45 & 0.50 & 0.58 \\
    \hline \hline
    \end{tabular}
    \caption{Prediction Errors on All Variables}
    \vspace{-1cm}
    \label{tab:prediction_table}
\end{table}

Fig.~\ref{fig:subjects_prediction} shows the predictions for the shank angle and the smart shoe heel pressure.
The red shaded area highlights the sensor readings, which have been used for conditioning, which generates the predictions starting at $0.25$ seconds and ending at $1.4$ seconds.
For both variables, the PIP predictions (blue) accurately match the ground truth values (black).
However, higher uncertainty in heel pressure persists during the swing phase, likely due to the variability in shoe pressure, resulting from air decompressing when lifting the heel.

Table~\ref{tab:comparison_table} shows how general the model is as well as a comparison to ProMP via. the errors present when tested with: PIP on the three training subjects, PIP two test subjects, and a ProMP model on two test subjects.

\subsection{\textbf{Experiment 2}: Latent Variable Estimation}
\begin{figure}[h!]
    \vspace{-.25cm}
	\begin{center}
		\includegraphics[width=1\columnwidth]{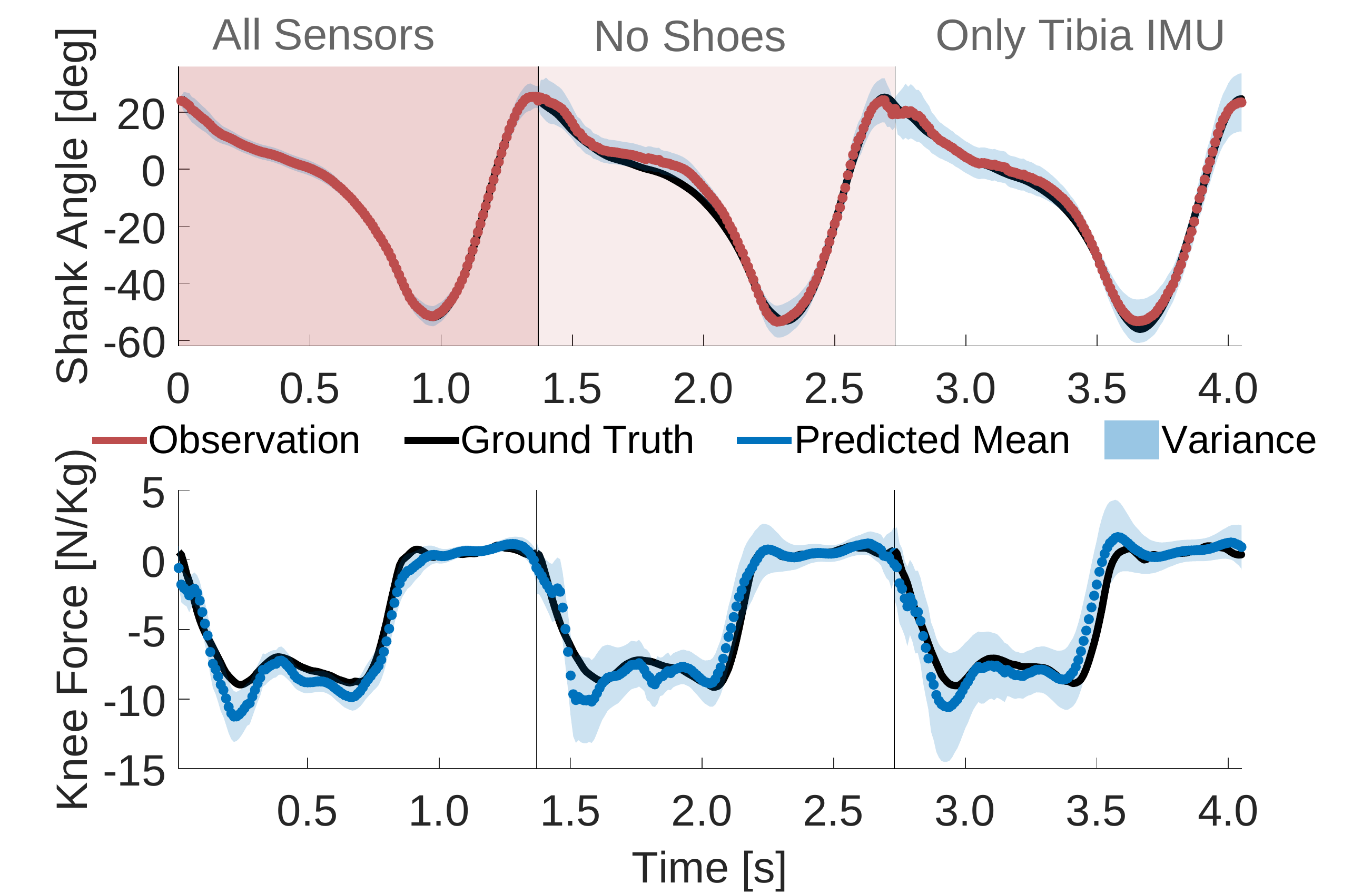}
	\end{center}
	\vspace{-.5cm}
	\caption{Inference of observable (top) and unobservable variables (bottom) when sensors start failing. Observations and inferences at each time step are shown as individual points. When sensor dropout occurs PIP does not fail catastrophically but rather, accuracy decreases gracefully thereby maintaining safety of the system.}
	\vspace{-.25cm}
	\label{fig:graceful_degradation}
\end{figure}

The second experiment aims at evaluating how accurately the method predicts latent \emph{unobservable} variables, e.g., biomechanical information that was added to the collected data during the data augmentation step (see Sec.~\ref{sec:augmentation}).
Fig.~\ref{fig:graceful_degradation} depicts the ground truth knee force along with the predicted values of the knee force as inferred via the PIP.
The bottom section of Table~(\ref{tab:prediction_table}) shows the kinematic and biomechanical variables that are inferred using this technique, namely ankle and knee: moments, forces and angles; along with the MAE for each subject. 
It is critical to note here, that the ground truth was obtained from the data augmentation process and therefore not composed of direct sensor measurements.
Still, we argue that by estimating these values PIP can provide helpful information needed to make informed decisions about the bodily ramifications of different types of motions. 

Besides prediction accuracy, a critical aspect when using machine learning components on an assistive device is \emph{safety}. It is important that predictions degrade gracefully (meaning minimal deterioration of the prediction accuracy) with noisy or missing sensor data.
Fig.~\ref{fig:graceful_degradation} shows an example for the graceful degradation as exhibited during inference in PIP, where sensors start failing one after the other.
In the first section of the figure, all sensors are observed when making predictions.
As sensors fail the uncertainty of the prediction grows and that the prediction accuracy slightly decreases. However, in general, the plot of the prediction trajectory still largely follows the ground truth. 

\begin{table}[]
\vspace{.25cm}
    \centering
    \begin{tabular}{lcccc}
    \multicolumn{5}{c}{\textbf{Predicted Variables MAE (Averaged Across Subjects)}}\\
    \hline \hline
         & & \thead{PIP Training} & \thead{PIP Test} & \thead{ProMP Test} \\
    \hline \hline
    Shank Angle & deg & 2.28 & 3.11 & 7.89 \\
    Shank Velocity & deg/s & 13.48 & 14.05 & 22.44 \\
    Femur Angle & deg & 1.59 & 1.58 & 6.77 \\
    Femur Velocity & deg/s & 6.62 & 6.88 & 13.45 \\
    Heel Pressure & mbar & 22.48 & 21.64 & 51.24 \\
    Meta 1 Pressure & mbar & 11.42 & 13.23 & 34.82 \\
    Meta 4 Pressure & mbar & 18.14 & 24.22 & 37.62 \\
    Toe Pressure & mbar & 23.30 & 23.21 & 55.63 \\
    Ankle Angle & deg & 1.34 & 2.21 & 10.10 \\
    Ankle Moment & N.mm/Kg & 13.54 & 17.66 & 25.45 \\
    Ankle Force & N/Kg & 0.15 & 0.14 & 0.31 \\
    Knee Angle & deg & 1.72 & 2.32 & 6.36 \\
    Knee Moment & N.mm/Kg & 43.13 & 56.61 & 76.48 \\
    Knee Force & N/Kg & 0.51 & 0.52 & 0.88 \\
    \hline \hline
    \end{tabular}
    \caption{Prediction Error Comparison of PIP and ProMP}
    \vspace{-1cm}
    \label{tab:comparison_table}
\end{table}

\subsection{\textbf{Experiment 3}: Learning Control with PIPs}
The last experiment evaluates the ability of PIPs to generate control signals for a powered lower limb prosthesis along with predictions. More specifically, we investigated if ankle angles from a healthy walking gait can be imitated and reproduced when wearing the active prosthesis using PIP.
Each participant was asked to wear the SpringActive Odyssey prosthesis~\cite{grimmer2016powered}, a battery powered lower limb prosthesis~\ref{fig:teaser}.
Imitation of the subjects walking gait is achieved by inferring in real-time the (unobservable) ankle angle and sending it to the prosthesis for execution.

\begin{figure}[t]
    \vspace{0.25cm}
	\begin{center}
		\includegraphics[width=1\columnwidth]{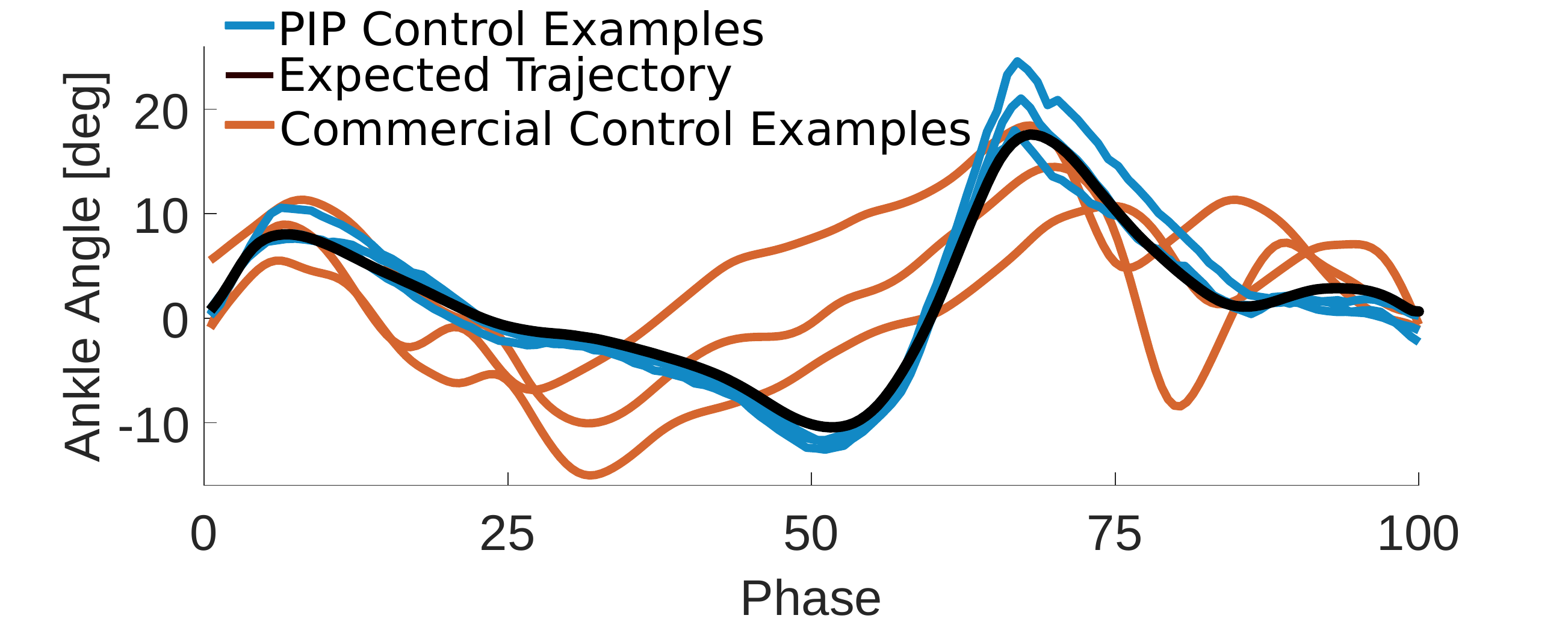}
	\end{center}
	\vspace{-.5cm}
	\caption{Ankle angle when no prosthetic was worn (black), when using the SpringActive control software (yellow), and when using PIP inference (blue).}
	\vspace{-.75cm}
	\label{fig:controls}
\end{figure}

Fig.~\ref{fig:controls} depicts the ankle angle that resulted from this experiment. The black line is an example of ankle angle when no prosthesis is worn. 
Ankle angles recorded when the prosthesis was used in conjunction with both PIP in blue and the commercial control software originally shipped with the SpringActive Odyssey in yellow, with three example strides of each.
It can be seen that while PIP closely tracks the expected ankle angle, the commercial control software substantially deviates from how the participant would naturally move.
A ProMP controller was not compared against in this experiment, since the high ankle angle error seen in Table~\ref{tab:comparison_table} would possibly make this controller unsafe.

\section{Conclusion}

In this paper, we proposed Periodic Interaction Primitives, a probabilistic framework that can be used to learn compact models of periodic behavior in symbiotic systems.
PIP facilitates prediction of future states of walking, as well as generation of control signals for assistive devices through its integrated phase estimation process.
Most notably this approach extends Interaction Primitives to latent variables which are either unobservable or hard to measure in practice.
For example, when collecting training data, we recorded motion capture readings to calculate offline biomechanical data such as ankle angles and internal forces of the subjects; but did not require additional motion capture readings at run-time.
Instead, we use other low-cost sensors to estimate the biomechanical variables at each time-step. 

Results in experiments with human participants indicate that Periodic Interaction Primitives efficiently generates predictions and ankle angle control signals for a robotic prosthetic ankle, even of subjects that the model was not trained on, with MAE of 2.21$^\circ$ in 0.0008s per inference.
Even on devices with limited compute power it is expected that inference rates of $1000$Hz can easily be achieved.
Compared to ProMP, with MAE of $10.10^\circ$ in 0.0169s per inference which is too slow and inaccurate for use with a robotic prosthetic. 
Furthermore it was shown that performance degrades gracefully in the presence of noise or sensor fall outs. 
This work shows that the approach can quickly generate predictive models and controls using a purely data-driven methodology.

\balance
\bibliographystyle{IEEEtran}
\bibliography{references}

\end{document}